\documentclass{article}

\usepackage{PRIMEarxiv}

\usepackage[utf8]{inputenc} 
\usepackage[T1]{fontenc}    
\usepackage{hyperref}       
\usepackage{url}            
\usepackage{booktabs}       
\usepackage{amsfonts}       
\usepackage{nicefrac}       
\usepackage{microtype}      
\usepackage{lipsum}
\usepackage{fancyhdr}       
\usepackage{graphicx}       
\usepackage{algorithm,algpseudocode}
\usepackage{adjustbox}
\usepackage{caption}

\graphicspath{{media/}}     

\title{Emulating the Human Mind: A Neural-symbolic Link Prediction Model with Fast and Slow Reasoning and Filtered Rules
}

\author{
  Mohammad Hossein Khojasteh \thanks{Corresponding author} \\
  School of Computer Engineering \\
  Iran University of Science and Technology \\
  Tehran, Iran\\
  \texttt{m\_khojaste@comp.iust.ac.ir} \\
   \And
  Najmeh Torabian \\
  School of Computer Engineering \\
  Iran University of Science and Technology \\
  Tehran, Iran\\
  \texttt{najmeh.torabian@gmail.com } \\
  \And 
  Ali Farjami \\
  School of Computer Engineering \\
  Iran University of Science and Technology \\
  Tehran, Iran \\
  \texttt{farjami110@gmail.com}
  \And 
  Saeid Hosseini \\
  Faculty of Computing and Information Technology \\
  Sohar University \\
  Sohar, Oman \\
  \texttt{saeid.hosseini@uq.net.au}
  \And 
  Behrouz Minaei-Bidgoli \\
  School of Computer Engineering \\
  Iran University of Science and Technology \\
  Tehran, Iran \\
  \texttt{b\_minaei@iust.ac.ir}
}

\begin{document}
\maketitle

\begin{abstract}
Link prediction is an important task in addressing the incompleteness problem of knowledge graphs (KG). Previous link prediction models suffer from \textbf{issues related to either performance or explanatory capability}. Furthermore, models that are capable of generating explanations, often struggle with \textbf{erroneous paths} or reasoning leading to the correct answer. To address these challenges, we introduce a novel Neural-Symbolic model named \textbf{FaSt-FLiP} (stands for Fast and Slow Thinking with Filtered rules for Link Prediction task), inspired by two distinct aspects of human cognition: "\textbf{commonsense reasoning}" and "\textbf{thinking, fast and slow.}" Our objective is to combine a logical and neural model for enhanced link prediction. To tackle the challenge of dealing with incorrect paths or rules generated by the logical model, we propose a semi-supervised method to convert rules into sentences. These sentences are then subjected to assessment and removal of incorrect rules using an NLI (Natural Language Inference) model. Our approach to combining logical and neural models involves first obtaining answers from both the logical and neural models. These answers are subsequently unified using an Inference Engine module, which has been realized through both algorithmic implementation and a novel neural model architecture. To validate the efficacy of our model, we conducted a series of experiments. The results demonstrate the superior performance of our model in both link prediction metrics and the generation of more reliable explanations.
\end{abstract}


\section{Introduction}
In recent years, knowledge graphs have received significant attention due to their suitability for data storage and their diverse applications, including question answering systems, chatbots, and recommendation systems. Within a knowledge graph, there are a vast number of entities and relationships, which together form facts. One important task directly related to knowledge graphs is link prediction. In this task, the objective is to deduce new facts based on existing knowledge within the graph. This task receives significant attention because, despite their size, knowledge graphs are incomplete. For instance, in one of the most famous knowledge graphs, Freebase, the birthplace information is absent for 71\% of individuals \cite{nguyen-2020-survey}.

The link prediction task has gained considerable attention due to its significance, leading to the development of various models to solve it. These models are categorized into three primary methods: embedding, logical, and multi-hop. Embedding models aim to transform entities and relationships into a vector space so that the graph structure is properly reflected in this vector space \cite{wang2014knowledge, dettmers2018convolutional}. Logical methods, on the other hand, typically involve the use of algorithms to extract a set of rules, often with associated confidence scores, which are then utilized to infer new facts \cite{meilicke2019anytime, ott2021safran}. Multi-hop models employ reinforcement learning to train models in a manner that enhances their ability to effectively navigate knowledge graphs and make predictions \cite{das2017go, lin-etal-2018-multi}.

Each of the link prediction model subgroups mentioned earlier has its own weaknesses. Embedding models can be classified into three subgroups: transitional distance models, tensor factorization, and neural models. Among these, neural models have demonstrated notably high accuracy. However, a major limitation of embedding models is their "\textbf{black box}" nature, lacking the ability to provide explanations for their generated output. In contrast to embedding models, the other two categories, namely logical and multi-hop models, tend to achieve \textbf{lower accuracies} (especially when compared to neural embedding models), but they possess the valuable capability to offer explanations for their output. Also, Our exploration of rules generated by the logical model has revealed that some of these rules are logically incorrect from a human perspective. Utilizing these incorrect rules to infer new triples leads to a \textbf{faulty reasoning process} and provides an unacceptable explanation for humans. The challenge of making inferences based on incorrect paths also exists in multi-hop systems \cite{lv-etal-2021-multi}.

Neural-Symbolic AI or Explainable AI is a group of models that try to use the advantages of neural models and symbolic models, i.e., high accuracy and explainability together \cite{sarker2021neuro}. In recent years, there has been a growing focus on these models for addressing various problems. Notably, for the link prediction task, neural-symbolic models have been proposed to tackle this challenge, with the RulE model standing out as an example \cite{tang2022rule}. In this article, we present a novel Neural-Symbolic model called \textbf{FaSt-FLiP} designed to address the link prediction problem. Our model draws inspiration from two key functions of the human brain: "\textbf{commonsense reasoning}" and "\textbf{thinking, fast and slow}." Commonsense reasoning represents a human ability to make assumptions or judgments in various life situations. For example, when presented with the statement "the cup dropped," an initial human inference might be "the cup is broken." This type of reasoning often obtains from spatial relations, causal relationships, scientific knowledge, and societal conventions encountered in daily life \cite{commonsensewiki2023, talmor-etal-2019-commonsenseqa}. The "thinking, fast and slow" theory, as proposed by Daniel Kahneman, posits that human thought processes are typically characterized by two distinct modes or systems. The first is "fast thinking," characterized by fast, automatic, and emotional reasoning, often relying on heuristics and shortcuts. The second is "slow thinking," which is more logical and analytical reasoning. Kahneman has elucidated the differences between these two systems and how they can produce contrasting results despite receiving the same inputs \cite{thinkingwiki2023}.

The main idea of this article is to develop an innovative Neural-Symbolic model by fusing a logical rules-based model with a neural model for the link prediction task. To tackle the \textbf{incorrect rules problem} generated by the logical model, we propose a solution: we initially convert rules into sentences using a hypothesis-premise format. Subsequently, we employ these sentences as inputs to a Natural Language Inference (NLI) model, which assigns a score to each rule. Any rule with a score below a defined threshold is identified as incorrect and is subsequently removed from the rule set. In our pursuit to create a Neural-Symbolic model that combines both logical and neural models, inspired by the "thinking, fast and slow" theory, we introduce a model that initially acquires answers from these two models. Subsequently, these generated answers are combined using an \textbf{Inference Engine} module, which is implemented via both an algorithm and a neural model. The algorithmic aspect of the Inference Engine receives answers from both the neural and logical models and combines them by employing the proposed algorithm. On the other hand, in the case of the neural inference engine, the scores from the logical model are integrated into the neural model during the early layers, effectively merging the neural model and the Inference Engine.

To assess the effectiveness of our proposed model, we conducted experiments using a well-known knowledge graph and evaluated its link prediction performance in comparison to models representing the three primary link prediction methods. The results of the experiments reveal that our model outperforms previous models in key metrics such as Hits@N while also providing more robust and informative explanations.

The contributions of our model can be summarized as follows: 

\begin{enumerate}
    \item We introduced a novel Neural-Symbolic model inspired by the "thinking, fast and slow" theory for the link prediction task.
    \item We presented a new algorithm and a novel architecture for creating the Inference Engine module, allowing the combination of results from a logical and neural model.
    \item A novel semi-supervised method for converting rules into sentences is proposed.
    \item We introduced an innovative approach to eliminating incorrect rules using Natural Language Inference (NLI) models.
    
\end{enumerate}

The rest of the paper is organized as follows. In Section \ref{sec:rel_works}, we present related works on the link prediction task and the NLI model. In Section \ref{sec:preliminaries}, we describe some preliminaries necessary for our model. Section \ref{sec:proposed_model} provides a detailed description of the proposed model. In Section \ref{sec:experiments}, we analyze the results of our model and compare them with previous link prediction models. \par

\section{Related Work}
\label{sec:rel_works}
The related work section begins by presenting an overview of primary link prediction methods, categorizing them into three main groups: Knowledge Graph Embedding, Multi-hop methods, and Logical and Rule-based Reasoning. Subsequently, the section explores hybrid approaches, focusing on Neural-Symbolic methods. Throughout the discussion, we will provide comparative analyses between the proposed model and the introduced models at the end of each subsection. To conclude this section, we will provide a brief explanation of NLI (Natural Language Inference) models and highlight their significance within the context of our proposed model.

\subsection{Link Prediction}
Despite their size, knowledge graphs often suffer from incompleteness. The link prediction task addresses this issue by inferring new triples from existing ones within the knowledge graph. Over the years, various models and methods have emerged for this purpose. Given a triple in the form $(h, r, t)$, either the starting or ending entity is omitted, creating a query in the format of $(h, r, ?)$ or $(?, r, t)$. Models are tasked with predicting the missing entity by leveraging the triples and information within the knowledge graph \cite{kumar2020link}.

\label{sec:knowledge-graph-embedding}
\subsubsection{Knowledge Graph Embedding}
\textbf{Knowledge graph (KG) embedding} represents a group of models aiming to comprehend the structure of a KG by transforming its components (entities and relations) into continuous vectors, facilitating various tasks like Link Prediction. These models can be categorized into three primary subgroups: a) \textbf{transitional distance models}, b) \textbf{semantic matching or tensor factorization-based models}, and c) \textbf{neural network-based models} \cite{dai2020survey}\cite{wang2017knowledge}. \par

In the translational distance model, a \textbf{distance-based scoring function} determines the validity of a fact based on the \textbf{embeddings} of two entities and the relation. TransE, the pioneering translational model, represents entities and relations as \textbf{low-dimensional vectors}. For instance, for a fact (h, r, t), the loss function computes the distance between the embeddings of h + r and t. The model's concept is rooted in the idea that the relation vector should serve as a link between two entities \cite{bordes2013translating}. Several subsequent models build upon TransE to address its limitations. For example, TransH enables entities to have distinct embeddings based on the involved relation \cite{wang2014knowledge}, while TransR extends this idea to different feature spaces for entities and relations \cite{lin2015learning}. Notably, TransD optimizes TransR by utilizing two vector embeddings for each entity-relation pair \cite{ji2015knowledge}. \par 

Tensor-factorization models, the second category of knowledge graph embeddings, create a \textbf{three-dimensional matrix} representing the knowledge graph and then \textbf{factorize} it based on a scoring function to derive embedding matrices. Each slice of the matrix signifies a unique relation, with elements set to one where connections between entities exist. Various tensor-factorization models employ different scoring functions to achieve improved factorization and, consequently, better embeddings. Models like RESCAL \cite{nickel2011three}, DistMult \cite{yang2014embedding}, ComplEx \cite{trouillon2016complex}, and RotateE \cite{sun2019rotate} fall into this category.

The surge in the popularity of \textbf{deep neural networks} has led to their utilization in embedding knowledge graphs into continuous vectors. Models like NTN \cite{socher2013reasoning}, MLP \cite{dong2014knowledge}, and NAM \cite{liu2016probabilistic} are among the early deep learning-based models for knowledge graph embeddings. ConvE, a notable model in this category, effectively embeds knowledge graphs by first concatenating the embeddings of the head entity and relation to create a matrix. A convolutional layer is then applied to generate multiple feature maps, which are subsequently projected into the embedding space and matched with the tail entity's embedding \cite{dettmers2018convolutional}.

Knowledge graph embedding models, while proficient at capturing KG structure, frequently suffer from a \textbf{lack of interpretability}. In our proposed model, we address this limitation by seamlessly integrating the outcomes of an embedding model with a logical model. This integration is achieved by incorporating the embedding model's outputs as inputs into our inference engine, making the embedding model an integral component of our proposed framework. Consequently, this combination of an embedding model and a logical model endows our final model with both interpretability and heightened accuracy.

\subsubsection{Multi-Hop Reasoning}
Multi-hop reasoning models constitute the second category of link prediction models applicable to knowledge graph completion. These models are rooted in \textbf{reinforcement learning} and are distinct from knowledge graph embedding models in their capacity to offer explanations. When presented with a query in the form of (h, r, ?), the primary objective of these models is to navigate the knowledge graph to identify the correct entity, with the path traversed serving as the model-generated explanation.

One of the pioneering models in employing Reinforcement Learning for multi-hop reasoning is MINERVA. The MINERVA model redefines the link prediction problem to align with a reinforcement learning framework. In this formulation, the state encompasses the combination of the query, answer, and current location, the action space encompasses all potential relations, and ultimately, a binary reward discerns whether the model successfully arrives at the correct answer \cite{das2017go}. Lin et al. enhanced MINERVA by modifying its binary reward mechanism using a pre-trained embedding model to mitigate the influence of false negative supervision. They also introduced action dropout to encourage the model to explore more diverse paths \cite{lin-etal-2018-multi}. RuleGuider is another model that revised MINERVA's binary reward. Utilizing high-quality rules, it furnishes reward guidance for the agent model \cite{lei-etal-2020-learning}. DIVINE, inspired by the generative adversarial concept, incorporates a discriminator to generate rewards and guide the RL agent \cite{li2019divine}. An intriguing approach by Hildebrandt et al. involves deploying two reinforcement learning models to engage in a debate over the correctness of a query \cite{hildebrandt2020reasoning}. Fu et al. also embraced the idea of two-agent models. In their design, the first agent is responsible for extracting relevant facts, while the second agent employs the facts recommended by the first agent for reasoning within the knowledge graph \cite{fu-etal-2019-collaborative}. Recently, Zhang et al. introduced another dual-agent model, with one agent navigating by clusters and the other by entities \cite{zhang2022learning}.

Both our proposed model and multi-hop models possess the capability to generate explanations for predicted links. Nevertheless, their methodologies exhibit notable distinctions. Multi-hop models primarily rely on reinforcement learning (RL) as their foundation for addressing the problem. In contrast, our approach integrates embedding and logical models within a hybrid framework. This hybrid model not only yields \textbf{superior performance} but also preserves the capacity to generate explanations for predictions. Furthermore, our model introduces a novel mechanism for eliminating unreasonable rules, enhancing the \textbf{reliability of the paths} it provides.

\subsubsection{Rule-Based Reasoning}
Rule-based reasoning models, akin to the previous category, offer the capability to generate interpretable outcomes for the Link Prediction task. However, in contrast to Multi-Hop models, their explanations encompass associated rules rather than paths. Some models, such as DRUM, referred to as neural-based, provide relevant rules while performing the task \cite{sadeghian2019drum}. In contrast, rule mining models extract rules and subsequently employ them for triple completion. Notably, AnyBurl stands out in this category, garnering substantial attention in recent years. AnyBurl introduces an algorithm named Anytime Bottom-up Rule Learning to extract rules. Following rule extraction, the Link Prediction task is executed using the Noisy-Or aggregation strategy applied to the acquired rules \cite{meilicke2019anytime}. SAFRAN further enhances AnyBurl by introducing a novel technique called Non-redundant Noisy-Or aggregation. This technique entails the detection and clustering of redundant rules before aggregation, resulting in improved performance \cite{ott2021safran}.

Similarly to the distinctions and commonalities observed between our model and Multi-hop models, our hybrid model shares the capacity for generating explanations with Logical models. However, the unique advantage of our hybrid model lies in its ability to attain \textbf{higher accuracy} and produce more \textbf{reasonable paths}.

\subsubsection{Neural-Symbolic models}
Symbolic neural artificial intelligence is a specialized branch of artificial intelligence aimed at \textbf{combining neural and symbolic models} to harness the strengths of both paradigms. Here, "neural model" refers to artificial neural networks known for their robust learning capabilities but limited interpretability. In contrast, the term "symbolic" pertains to AI models, such as logical models, that manipulate symbols that can offer explanations for their outputs. While symbolic models may not match the learning prowess of neural models, they excel in providing interpretability. The combination of these models serves the dual purpose of enhancing model accuracy while retaining the ability to have explanations \cite{sarker2021neuro}.

In recent years, the field of neural-symbolic models has garnered significant attention, with some experts considering it the next generation of AI model development. Within the domain of link prediction, several neural-symbolic models have emerged. For instance, Niu et al. introduced a four-stage closed-loop framework designed to fuse neural and symbolic reasoning. In their model, rules, akin to relations in the TransE model, are embedded, serving as transition vectors between the embeddings of head and tail entities in positive triples \cite{niu2021perform}. Another notable model in this realm is RulE, which posits that each rule should possess an embedding, with this embedding serving as a transition between the relations inherent to the rule. To acquire embeddings for individual rules, they introduced a novel score function, incorporating negative sampling techniques tailored for rules. Following the training phase, a score can be computed for each rule based on this score function, subsequently applied in the context of soft rule reasoning \cite{tang2022rule}.

Two fundamental distinctions set our proposed model apart from the introduced neural-symbolic models. Firstly, in our model, we subject the extracted rules to evaluation by a Natural Language Inference (NLI) model, allowing us to identify and discard rules deemed unreasonable. This approach aligns with human reasoning, where the acceptance of rules is often based on experience rather than relying solely on embeddings for scoring. Secondly, we advocate for a dual-path reasoning approach, segregating the logical and neural models into distinct paths. Subsequently, we integrate their respective results utilizing an inference engine. This approach mirrors human reasoning patterns, where individuals employ different paths for reasoning and then amalgamate the outcomes, giving precedence to the significance of each path. Thus, our model endeavors to emulate this cognitive process by facilitating neural and logical reasoning paths, followed by the synthesis of their results through an inference engine.

\subsection{Natural Language Inference (NLI) models}
Natural language inference (NLI) constitutes a task within the field of Natural Language Processing (NLP). It is essentially a 3-way classification problem, categorizing statements into three labels: entailment, contradiction, and neutral. The primary objective is to determine whether the provided premise logically implies a given hypothesis. To elucidate, entailment signifies a conclusion that a person would typically infer from the premise or the implied information conveyed by the premise. Contradiction, on the other hand, denotes that the hypothesis contradicts the information provided by the premise. Lastly, neutral indicates that there is insufficient information to make a conclusive determination regarding the hypothesis \cite{yu2023nature}. The SNLI \cite{bowman2015large} dataset and the MNLI \cite{williams2017broad} dataset stand out as renowned datasets for this particular task.

Transformer-based models, such as RoBERTa \cite{liu2019roberta} and DeBERTa \cite{he2020deberta}, typically serve as the preferred choices for this task. Initially, these models undergo training on fundamental tasks, like MASK Language, before being fine-tuned on the aforementioned datasets. Additionally, research efforts have been dedicated to addressing the limitations of these models in the context of the NLI task. For instance, Glockner et al. introduced a new test set that highlighted the poor performance of these models when confronted with challenges related to lexical and world knowledge \cite{glockner2018breaking}. In this article, we have employed these models to detect unreasonable rules extracted by a logical model.

\section{Preliminaries}
\label{sec:preliminaries}
A Knowledge Graph (KG) is a structured dataset comprising nodes and directed labeled edges connecting these nodes. In this representation, nodes correspond to entities existing in the real world, encompassing a wide array of subjects such as humans, cities, and more. Meanwhile, the edges denote relations between these entities, including descriptors like "place\_of\_birth" or "contain\_city." The set of entities is denoted as E, and the set of relations is denoted as R. The knowledge contained within a Knowledge Graph is expressed through triples, adhering to the format (h, r, t), where h and t are entities drawn from the set E, and r is a relation from the set R. For instance, the triple (Jack, place\_of\_birth, USA) encapsulates a piece of information within a KG.

\begin{equation}
KG = \{(h, r, t) | h, t \in E, r \in R\}
\end{equation}

Despite the vast expanse of Knowledge Graphs, they invariably suffer from incompleteness. To address this issue, the task of Link Prediction emerges, aimed at augmenting KGs by leveraging existing knowledge. In the context of Link Prediction, queries assume the format (h, r, ?), with the primary objective being to discern the most suitable entity to complete the given query. This determination relies on the available knowledge in other triples within the KG.

\subsection{Embedding-based reasoning}

As elucidated in Section \ref{sec:knowledge-graph-embedding}, the core concept behind knowledge graph embedding revolves around the transformation of entities and relations into continuous vectors via a scoring function. Subsequently, these learned embeddings, in conjunction with the score function, facilitate the retrieval of answers to queries posed within the knowledge graph. In our research, we have chosen the ConvE model \cite{dettmers2018convolutional} to serve as the embedding model component within our Neural-Symbolic framework.

\textbf{ConvE} \cite{dettmers2018convolutional}, a neural Knowledge Graph Embedding (KGE) model, has demonstrated its competitive prowess in the area of Link Prediction tasks. When confronted with queries in the format of (h, r, ?), ConvE's operational framework unfolds in several key stages. Initially, the model concatenates the embeddings of the head entity (h) and the relation entity (r), creating a matrix akin to an image. Subsequently, a convolutional layer is applied to this matrix, extracting a set of feature maps. These feature maps are then flattened and fed into a fully connected layer equipped with neurons mirroring the dimension of entity embeddings. The output of this fully connected layer essentially represents an embedding, intended to approximate the embedding of the tail entity. To gauge this proximity, the output embedding is multiplied by the embeddings of all entities, generating logits that reflect the degree of closeness to each entity. These logits are then subjected to a softmax layer, yielding probabilities for each entity as a potential correct answer. By comparing these probabilities to the actual correct answer, the network computes its loss. The ConvE model employs a specific score function to underpin its operations, as delineated below:

\begin{equation}
\label{equ:conve}
f(vec(f([\overline{e_{h}}; \overline{r_r}] * \omega)) W)e_t
\end{equation}

Where $e_h$ and $r_r$ and $e_t$ are the embeddings of the head entity and the relation and the tail entity respectively. Also, $\overline{e_{h}}$ and $\overline{r_r}$ are 2D reshaping of $e_h$ and $r_r$. We opted to integrate the ConvE model into our embedding-based reasoning module for two compelling reasons: its track record of yielding impressive results and its exceptional computational efficiency. ConvE has consistently demonstrated its competitive performance across various Knowledge Graphs (KGs). Moreover, a detailed analysis has revealed ConvE's efficacy in modeling nodes with high indegree, a common occurrence in KGs like Freebase. This capability is instrumental in enhancing the model's overall performance and utility. Furthermore, the utilization of a convolutional layer and a 1-N scoring procedure in ConvE leads to a significant reduction in the number of parameters compared to models such as DistMult \cite{yang2014embedding} (by a factor of 8x). This streamlined architecture not only expedites the training process but also drastically accelerates the evaluation phase, rendering it approximately 300 times faster.

\subsection{Rule-based reasoning}
\textbf{AnyBurl} \cite{meilicke2019anytime} represents a logical model designed to extract logical rules and subsequently employ these rules to enhance the Link Prediction task. The rules extracted by this model adhere to the format $r_0(h_0, t_0) \leftarrow r_1(h_0, t_1), ..., rn(h_{n-1}, t_0)$. This rule structure signifies that if the triples $(h_0, r_1, t_1), ..., (h_{n-1}, r_n, t_0)$ are present within the knowledge graph, then it is reasonable to deduce that the triple $(h_0, r_0, t_0)$ is a valid addition to the KG. For instance, consider this rule:

\begin{equation}
\label{eqn:examplerule}
Speak(X, Y) \leftarrow Lives(X, A), Language(A, Y)
\end{equation}

This rule implies that if an individual X resides in a country A, and the official language of country A is Y, then one can logically infer that person X is capable of speaking language Y. In this context, the right part of the rule is referred to as the rule body or premise, while the left part is designated as the rule head or conclusion. It's noteworthy that the size of a rule is determined by the number of atoms within the rule body. For instance, the rule \ref{eqn:examplerule} has a size of 2. Importantly, rules extracted by the AnyBurl model adhere to a maximum size of 3. An intriguing observation is that Multi-hop models, such as MINERVA, similarly employ paths with a maximum length of 3 \cite{das2017go}.

To derive logical rules from Knowledge Graphs (KGs), AnyBurl employs an algorithm known as "Anytime Bottom-Up Rule Learning." This algorithm entails conducting a random walk from the head entity to the tail entity, subsequently extracting a rule based on the traversed path. After that, a score is computed for the rule, and if it surpasses a predefined threshold, the rule is retained; otherwise, it is discarded. Within the framework of the AnyBurl model, a confidence score is computed for each rule. The formula for calculating this score is as follows:

\begin{equation}
\label{eqn:confidence}
Confidence = \frac{Correctly Predicated}{Predicated}
\end{equation}

As depicted in Equation \ref{eqn:confidence}, the confidence score assigned to each rule is computed by dividing the number of correct predictions made by the rule by the total number of predictions. This score serves as a metric to gauge the correctness of a rule, taking into account the KG's structure and existing knowledge.

Once a set of rules with their associated scores is extracted, they can be leveraged for the Link Prediction task. When presented with a query in the format of (h, r, ?), the process unfolds as follows: First, rules are extracted in which the relation r appears in the rule head or conclusion. Subsequently, the head entity h is employed as input to all the extracted rules, utilizing forward chaining to compute all the possible tail entities that may serve as answers. For each tail entity generated by the rules, a score is computed to quantify the value of each potential answer. In the AnyBurl model, an entity's score is equivalent to the highest confidence score among the rules capable of deducing that entity. In cases where two entities possess identical scores, the second-best confidence score is utilized as a tiebreaker. This process continues until all entities have distinct scores. Subsequently, the sorted list of potential answers is harnessed to compute performance metrics such as Hits@N. It is worth noting that this article introduces a slight modification to the method for calculating entity scores, which will be elaborated upon in the subsequent section.

\textbf{Problem with Entities in Rules: }In this article, we employed the result rules extracted by the AnyBurl model. Upon acquiring the rule set, we encountered two significant issues within the rules. The initial concern pertained to rules containing entities, illustrated by the following examples that has extracted from the freebase knowledge graph:

\begin{equation}
\label{eqn:entrules1}
nationality(X,/m/02jx1) \leftarrow languages(X,/m/02h40lc)
\end{equation}

\begin{equation}
\label{eqn:entrules2}
nationality(X,/m/02jx1) \leftarrow languages(X,/m/064\_8sq)  
\end{equation}

As illustrated, these rules serve the purpose of determining an individual's nationality based on their spoken language. The underlying concept is sound, as different languages are typically associated with specific countries, enabling the inference of a person's nationality based on their language proficiency. The extracted rules presented above are specific to a particular nationality $(/m/02jx1)$. Based on these rules, it is evident that the languages $/m/02h40lc$ and $/m/064\_8sq$ are spoken in the country $/m/02jx1$. Consequently, if an individual can converse in these two languages, it can be deduced that their nationality aligns with the country $/m/02jx1$.

However, a notable drawback of these rules lies in their failure to establish a reasoning path; instead, they rely on external knowledge gleaned from the Knowledge Graph (KG). In contrast, consider Rule \ref{eqn:examplerule}, which proposes a clear reasoning path and does not pertain to a specific entity. Rules \ref{eqn:entrules1} and \ref{eqn:entrules2}, on the other hand, are tied to a particular country and two languages, offering knowledge extracted from the KG rather than a logical reasoning path.

A second issue with these types of rules arises from the potential for contradictions among different rules. For instance, while it is a known fact that each country has a designated official language, Rules \ref{eqn:entrules1} and \ref{eqn:entrules2} suggest two languages for a single country. Consequently, only one of these languages can be the official language, while the other is simply a frequently spoken language within that country. For example, although French is spoken in various cities in Canada, deducing an individual's Canadian nationality solely based on their ability to speak French is untenable.

Due to these inherent limitations, characterized by the absence of a clear reasoning path and the potential for contradictions, we made the decision to eliminate rules containing entities. This action resulted in a reduction in the number of rules, decreasing the total from approximately 2 million to approximately 53,000 rules.

\label{sec:incorret_rules}
\textbf{Problem with Incorrect Rules: }It is imperative to acknowledge that rules extracted from knowledge graphs primarily serve as indications that the rule head can be inferred from the rule body; they do not necessarily denote causality between them. To illustrate, in the quest to deduce a movie's director, one might initially extract the genre of that movie and subsequently propose directors who specialize in that genre as potential candidates. The critical criterion is that these suggestions must align with human reasoning and be deemed acceptable explanations. Upon scrutinizing the extracted rules, it became evident that certain rules suggested explanations that did not align with human reasoning. Notably, some of these erroneous suggestions carried high confidence scores. This issue of incorporating unreasonable paths is not unique to our model; it also plagues Multi-Hop methods, as elucidated by Lv et al \cite{lv-etal-2021-multi}. 
An illustrative example provided by Lv et al. involves attempting to deduce a person's cause of death solely based on the location of their death. This example highlights an incorrect and unacceptable explanation and reasoning path.

For instance, consider the following rules:

\begin{equation}
\label{eqn:wrongrule1}
film/produced\_by(X,Y) \leftarrow film/edited\_by(X,A), person/sibling(A,Y)  
\end{equation}

\begin{equation}
\label{eqn:wrongrule2}
person/gender(X,Y) \leftarrow film/written\_by(A,X),film/music(A,B), person/gender(B,Y)
\end{equation}

\begin{equation}
\label{eqn:wrongrule3}
ethnicity/distribution(X,Y) \leftarrow ethnicity/distribution(X,A), location/exported\_to(A,Y)     
\end{equation}

In the aforementioned examples, we present instances of rules whose suggestions are deemed unacceptable by human standards. For instance, Rule \ref{eqn:wrongrule1} posits that the sibling of a movie's editor is the producer of said movie, a premise that is evidently incorrect. This rule, surprisingly, carries a confidence score of approximately 0.96. Similarly, Rule \ref{eqn:wrongrule2} attempts to deduce that the writer and musician of a movie share the same gender, which is plainly erroneous. Lastly, Rule \ref{eqn:wrongrule3} seeks to infer the geographic distribution of an ethnicity, asserting that if a particular ethnicity is present in location A and location A has exportation to location Y, then the ethnicity must also be present in location Y. This assertion, however, runs counter to real-world scenarios where various ethnicities from different countries engage in exportation relations.

One of the primary objectives of this article is to identify and subsequently eliminate such erroneous rules through the utilization of the Natural Language Inference (NLI) model, an endeavor that will be expounded upon in the ensuing section.

\section{Method}
\label{sec:proposed_model}
\begin{figure}
  \centering
  \includegraphics[width=1.0\textwidth,height=8cm]{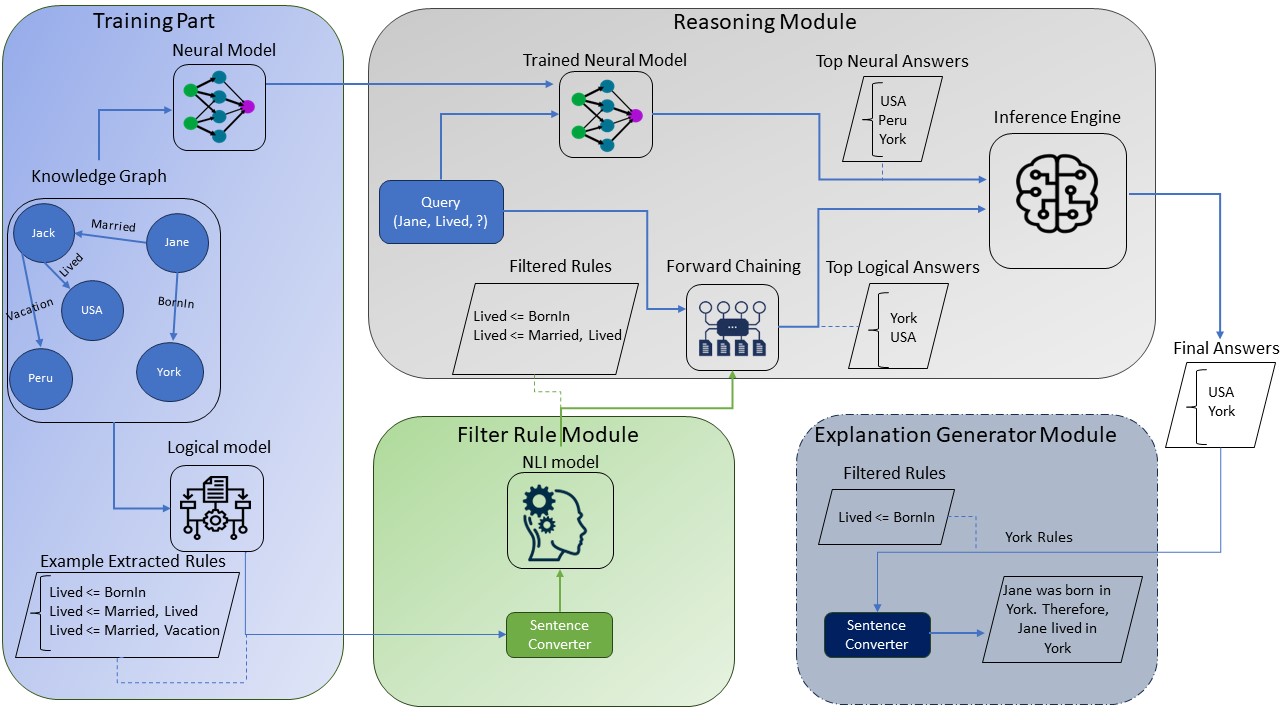}
  \caption{Our proposed FaSt-FLiP model. a) Training Part: This initial phase involves two crucial steps. First, the neural model is meticulously trained, equipping it with the ability to process and interpret knowledge graph data. Simultaneously, the logical model extracts logical rules from the knowledge graph. b) Filter Rule Module: Within this module, the identification and removal of unreasonable rules is done. This task is achieved through the collaboration of two sub-modules. The Sentence Convertor takes these rules and transforms them into coherent and human-readable sentences. Following this, a Natural Language Inference (NLI) model receives these sentences as input and assigns scores to evaluate the correctness of each rule. c) Reasoning Module: The heart of our model lies in this module, where the results from the neural and logical models converge. A pivotal component known as the Inference Engine takes charge of combining these results. This engine can function as either an algorithm or a neural model. d) Explanation Generation Module: The final piece of the puzzle focuses on enhancing the user experience. This module is dedicated to crafting explanations in natural language for the answers generated by the model. It accomplishes this by drawing upon the extracted rules and employing the Sentence Convertor to translate them into human-understandable explanations.}
  \label{fig:fig1}
\end{figure}

In this section, we will delve into the details of our model named "FaSt-FLiP" (Fast and Slow Thinking with Filtered rules for Link Prediction task). The overall structure of our proposed model is visually represented in Figure \ref{fig:fig1}. This diagram delineates four distinct modules or phases within the model. The primary objective of our proposed model is to combine the outcomes of two distinct reasoning paths: embedding and logical rule-based. To enhance the logical model's credibility, we introduce a process for eliminating unreasonable rules extracted by the logical model through the utilization of Natural Language Inference (NLI) models.

The initial part of our proposed model is the training phase. During this phase, the neural model undergoes training, and logical rules are extracted from the Knowledge Graph (KG) based on a predefined algorithm. Importantly, this training phase can be conducted independently, and for our proposed model, we simply require the extracted rules and the trained neural model.

Within the first module of our proposed model, we address the identification and removal of unreasonable rules, as elaborated upon in the previous section. Unreasonable rules are those that a human observer would deem unacceptable as explanations. To tackle this challenge, we transform each rule into two sentences in the premise-hypothesis format and subsequently employ an NLI model to assign a score to each rule. Any rule that falls below a predetermined threshold score is eliminated.

The reasoning module assumes the responsibility of merging the outcomes of both the neural model and the logical model. Given a query, we extract the sorted results of answers from the neural model alongside the filtered rules from the logical model. These results, complete with their respective scores, are fed into an inference engine tasked with consolidating the results to yield the final outcomes. The inference engine can take the form of an algorithm or even a neural model designed for result fusion. Finally, for each of the answers obtained, we have the capability to generate natural language explanations based on the rules within the Explanation Generator module. These explanations enhance the interpretability and transparency of the model's decisions.

\subsection{Filter Rule Module}

The Filter Rule Module serves the primary purpose of identifying and eliminating unreasonable rules from the rule set. As previously discussed in Section \ref{sec:incorret_rules}, rules extracted from the knowledge graph merely imply that the rule head can be inferred from the rule body, without implying a causal relationship between the head and body of the rule. In the preceding section, we presented examples of unreasonable rules that would not be deemed acceptable as valid reasoning paths. In the subsequent section, we will outline our methodology for the detection and removal of these unreasonable rules. 
As depicted in Figure \ref{fig:fig1}, this process can be divided into two main steps: converting rules into sentences and utilizing these sentences as input for an NLI (Natural Language Inference) model.

\textbf{Sentence Convertor:} The rules extracted from the Knowledge Graph (KG) are structured as $x \leftarrow y$, with $x$ representing the rule head or conclusion and $y$ constituting the rule body or premise. Our objective is to devise a method that allows us to express the probability of a rule's correctness, namely, how likely it is for $x$ to be inferred from the rule body $y$. Remarkably, the format of these rules aligns with that of the Natural Language Inference (NLI) task, which involves a premise and a hypothesis, with the aim of determining whether the hypothesis can be deduced from the premise. In the context of these rules, the rule head corresponds to the hypothesis, and the rule body serves as the premise. Thus, we approach the assessment of rule correctness as an NLI task, where the rule head functions as the hypothesis, and the rule body serves as the premise.

As mentioned earlier, the problem of identifying unreasonable rules can indeed be reframed as a Natural Language Inference (NLI) task. However, a key challenge arises when applying NLI models since they require input in the form of natural language text. Consequently, rules cannot be directly fed into these models; instead, they must be transformed into sentences. One approach to converting rules into sentences is to adopt the concept of Machine Translation, where rules and sentences are treated as two distinct languages. However, a significant challenge with this approach lies in the absence of a dataset containing sentence-rule pairs. Creating such a dataset manually is a challenging and time-consuming task. Consequently, we have employed a semi-supervised method to facilitate the conversion of rules into sentences.

The fundamental approach to converting rules into sentences involves two key steps. Initially, we transform the relations within the rules into sample sentences. Subsequently, we combine the sentences related to each rule's relations to generate a complete sentence representing the rule. To illustrate this process, consider the following rule: $person/nationality(X,Y) \leftarrow person/place\_of\_birth(X,Y)$

Initially, we convert each relation into sentences. For instance, we transform the relation $person/nationality$ into the sentence "June's nationality is American" and the relation $person/place\_of\_birth$ into the sentence "Jack's place of birth is Honolulu, Hawaii." To form a complete sentence representing the rule, we combine the sentences associated with each relation. In this example, the combined sentence reads: "Jack's place of birth is America. Therefore, Jack's nationality is American." Notably, in this process, the names of the individual and the country are synchronized to ensure a coherent sentence structure.

The primary advantage of the proposed approach for converting rules into sentences lies in its simplicity and efficiency. This method leverages the fact that the number of relations is significantly smaller than the number of rules (237 relations compared to 53,000 rules). Consequently, converting relations into sample sentences is a more efficient and optimal strategy than directly converting the rules. As previously outlined, this approach involves two main stages: 1) converting relations into sample sentences and 2) combining the sentences associated with the relations to construct complete sentences for each rule.

To convert relations into sentences, we employed a semi-supervised method utilizing ChatGPT-3. This approach involved soliciting ChatGPT to transform each relation into a meaningful sample sentence. ChatGPT was employed to expedite the conversion of relations, although a human could perform this entire process. Upon obtaining the results from ChatGPT, a human evaluator reviewed all the outputs and made adjustments to the sentences in cases where issues arose. Ultimately, with the combined efforts of ChatGPT and a human evaluator, we extracted the most likely types of head and tail entities, a sample sentence, and the positions of entities within the sample sentence for each relation. The entity types and their positions were essential for aligning the relations of the rules, a topic that will be discussed shortly. For example, the information extracted for the relation $/people/person/nationality$ included:

\begin{itemize}
\item Type of entities: Head Entity: Person, Tail Entity: Country
\item Sample sentence with the position of entities: [Samantha Smith]'s nationality is [American].
\end{itemize}

Combining the sentences derived from relations to generate a final sentence for each rule entails more than simply concatenating these sentences in sequence. The primary challenge in this process lies in aligning the entities within the relations to produce a meaningful and accurate sentence. Some instances are provided in the following:
\begin{itemize}
\item If a relation pertains to nationality, the entity should represent a country name, not a city name, and preceding or subsequent relations should employ this country name consistently.
\item When two relations share the same entity, it must bear the same name in the final sentence (e.g., "Jack's place of birth is America. Therefore, Jack's nationality is American" is correct, while "Jack's place of birth is America. Therefore, Jane's nationality is American" is incorrect).
\item If the relations involve distinct entities, they should not share the same name in the final sentence (e.g., "America exported to America" is an incorrect sentence).
\end{itemize}

We have established a set of general conditions (as demonstrated in the second and third examples) and specific conditions unique to each relation (as illustrated in the first example). These conditions were identified through the examination of multiple rules and have been implemented in Python code. Consequently, this Python code functions as a Sentence Converter module, taking a rule as input and yielding a natural language sentence as output.

In summary, the process of converting rules into sentences involves two key steps: 1) The conversion of relations into sentences is accomplished by employing ChatGPT in conjunction with human oversight. This phase results in the creation of meaningful sample sentences for each relation. 2) The sentences derived from relations are then combined using a Python code that incorporates various conditions for ensuring entity synchronization. This code functions as a Sentence Converter module, generating natural language sentences for each rule. This process raises two primary questions for consideration: 1) What are the advantages of this automated process compared to having a human observer manually convert all rules into sentences? 2) Can a sample sentence effectively serve as a substitute for the rule itself in assessing the correctness of that rule?

To address the first question, we can highlight several advantages of the implemented process over manual conversion by a human:
\begin{itemize}
\item \textbf{Efficiency}: The process is significantly more efficient as the number of relations is approximately 223 times smaller than the number of rules. This means a substantial reduction in the workload. Additionally, the workload was further reduced by leveraging ChatGPT.
\item \textbf{Ease of Rule Combination}: Extracting the conditions needed to combine rules is relatively straightforward and can be determined by examining a few rules. This simplifies the process.
\item \textbf{Reduced Cognitive Load}: Converting rules to sentences can be a challenging task for a human as it requires careful reading of rules, comprehension of rule concepts, and an understanding of entity positions.
\item \textbf{Scalability}: The implemented module can easily handle the conversion of new rules into sentences without the need for human labeling, making it scalable for larger datasets with similar relations or new rules.
\item \textbf{Potential for Dataset Creation}: The sentence-rule pairs generated by this process can serve as a foundation for creating a dataset for translating rules into sentences, which could be beneficial for future research.
\end{itemize}

In this article, the Sentence Converter module's output is sufficient for our purposes, and there was no immediate need to create a dataset and train a model for rule-sentence translation.

The answer to the second question is that sample sentences can indeed serve as substitutes for rules. This is because alterations in sentences can be achieved by modifying the entities within the sentence, without altering the core idea of the sentence. For instance, the sentence "Jack’s place of birth is America. Therefore, Jack’s nationality is American." can be transformed into the sentence "June's place of birth is England. Therefore, June’s nationality is English." These two sentences convey the same concept, and changing the entities does not impact the meaning or the reasoning pathway of the sentence. Consequently, sample sentences can effectively represent the reasoning pathways established by rules, making it possible to use them to assess the correctness of the rules.

\textbf{Using NLI Models on Rules:} Once sentences have been generated for each rule, they can be employed as input for an NLI (Natural Language Inference) model to assess the correctness of the rules. To facilitate this process, we initially fine-tuned the NLI model using a small random subset of rules, comprising approximately 10\%. Subsequently, we applied the NLI model to the rules to obtain probabilities associated with three possible outputs: entailment, neutral, and contradiction. The final score for each rule was computed as follows:

\begin{equation}
\label{eqn:nliscore}
Final_{score} = Entailment_{score} + \gamma * Confidence * Neutral_{score}
\end{equation}

In Equation \ref{eqn:nliscore}, the $Entailment_{score}$ and $Neutral_{score}$ scores represent the probabilities of entailment and neutrality, respectively, obtained from the NLI model. $Confidence$ is the score derived from the knowledge graph, as detailed in Equation \ref{eqn:confidence}. Lastly, $\gamma$ serves as a hyperparameter, indicating the relative importance of the two components within this equation. The underlying idea behind this formula is to gauge the correctness of a rule by assessing the likelihood of not having a contradiction between the rule head and body. If a rule exhibits a high probability of entailment, it suggests that the rule could potentially be correct. A rule with a high probability of neutrality will have its final score determined by the confidence score from the knowledge graph. Conversely, if the contradiction score is high, the final score will be low. Consequently, rules with final scores surpassing a specified threshold are retained in the rule set, while those failing to meet the threshold are removed. The formula for this process is outlined below:
\begin{equation}
\label{eqn:finalscore}
If \; Final_{score} > th \; then \; return \; Confidence \; else \; return \; 0.
\end{equation}

\subsection{Reasoning Module}
In this section, the reasoning component of the proposed model will be explanied. As illustrated in Figure \ref{fig:fig1}, part C, the central objective of the Reasoning Module is to amalgamate the outcomes of two distinct models. These models are founded on different concepts, with one striving to tackle the problem by learning embeddings for entities and relations, while the other focuses on rule extraction and utilizes these rules to address the problem. The crux of developing a hybrid Neural-Symbolic model for the Link Prediction task hinges on the harmonious integration of these two models. A pivotal distinction between our model and previous Neural-Symbolic models lies in the fact that our neural and logical models conduct their reasoning independently. Subsequently, the results are unified through an Inference Engine. To showcase the effectiveness of the Inference Engine, this module has been implemented both as an algorithm and a neural model, which we will elucidate shortly.

In our Reasoning Module, we employed two models: ConvE as the neural model and AnyBurl as the logical model. Both of these models were discussed in detail in Section \ref{sec:preliminaries}. One minor adjustment made to the AnyBurl model involves a slight modification to the score function. The formula for this updated score function is as follows:

\begin{equation}
\label{eqn:anyburlscore}
ent_{score} = \Sigma_{i=1}^{7} \frac{S_i}{100^{i-1}}
\end{equation}

Formula \ref{eqn:anyburlscore} was employed to calculate the score for each entity in the logical model. For a given query, all rules that match the relation in the query are utilized to infer the entity answers. These rules are then sorted based on their confidence score in a set for each entity. In Formula \ref{eqn:anyburlscore}, $S_i$ represents the score of the ith rule in the sorted list of rules capable of inferring the entity. Each score is divided by $100^{i - 1}$. This normalization is applied to ensure that an entity inferred by a single rule with a score of 0.7 receives a higher score than an entity inferred by two rules with scores of 0.6 and 0.5. Additionally, to maintain computational efficiency, we limited the number of rules considered to 7, as experiments revealed that beyond this threshold, rules typically have very low scores, and including them does not significantly impact the entity score calculation.

\textbf{Using Algorithm for Inference Engine }In the initial part of this section, we will elucidate the implementation of the Inference Engine using an algorithm. Utilizing an algorithm for combining results offers the advantage of full explainability, albeit at the cost of lower accuracy compared to using a neural model. The algorithm's implementation is detailed in Algorithm \ref{alg:inferenceengine}.

\begin{algorithm}
\caption{Inference Engine Algorithm}
\label{alg:inferenceengine}
\textbf{Input: } Conve Entity Answers dictionary for query $A = \{e_{11}: s_{11}, e_{12}: s_{12}, ..., e_{1n}: s_{1n}\}$; Anyburl Entity Answers dictionary for query $B = \{e_{21}: s_{21}, e_{22}: s_{22}, ..., e_{2n}: s_{2n}\}$; boolean Flag.\Comment{The Entity Answers dictionaries are the outputs of ConvE and Anyburl models with their scores.}\par
\begin{algorithmic}[1]
\State \textbf{Initialize} Final Entity Answers C to an empty array, Empty Binary Tree T.
\For{$ent$ in $A$}
    \State \textbf{Set} score to zeros
    \If{$Flag = 0$} 
        \If{$ent$ exist in $B$}
            \State \textbf{Set} score to $A[ent]$
        \Else 
            \State \textbf{Set} score to -1
        \EndIf
    \Else 
        \If{$ent$ exist in $B$}
            \State \textbf{Set} score to $A[ent] + B[ent]$
        \Else 
            \State \textbf{Set} score to $A[ent]$

        \EndIf
    \EndIf
    \State \textbf{Add} $ent$ with score to $T$ \Comment{Sort will be done}
\EndFor

\If{$Flag = 1$}
    \For{$ent$ in $B$}
        \If{$ent$ not exist in $A$}
            \State \textbf{Add} $ent$ with $B[ent]$ to $T$
        \EndIf
    \EndFor
\EndIf

\State \textbf{Set} $C$ to flatten $T$ 

\State \Return $C$
\end{algorithmic}
\end{algorithm}

Algorithm \ref{alg:inferenceengine} takes as input the results from both the neural model and the logical model. These results consist of entities that could potentially be answers for a query in the format of (e1, r, ?), along with their corresponding scores. Additionally, a boolean parameter named "flag" is provided as input, serving as a boolean parameter that determines the method for combining the results from both models.

Two distinct approaches are available for result combination: a) Eliminating entities from the answer set of the neural model if they cannot be justified by a path from the logical model, while preserving the original order generated by the neural model, and b) Aggregating the scores from both the neural model and the logical model, followed by sorting to generate the final list of entities. The choice between these two approaches is controlled by the boolean "flag" parameter, which is a hyperparameter based on the validation set for each relation.

The algorithm employs a binary tree for adding and sorting entities efficiently. For each entity, the algorithm calculates its score based on the flag parameter and the results obtained from both models. The calculated score is then inserted into the binary tree, which is sorted during the insertion process. Once scores for all entities have been computed, the binary tree is flattened and returned as the final answer.

To estimate the time complexity of the algorithm, let $n$ denote the number of result answers. For each of these $n$ entities, the algorithm calculates the score and adds it to the binary tree. Adding and sorting in a binary tree has a time complexity of $O(log n)$ in the worst case. Additionally, dictionary lookups in hash tables (dictionaries A and B) have a time complexity of $O(1)$. Consequently, the overall complexity of the algorithm is $O(n log n)$.

\begin{figure}
  \centering
  \includegraphics[width=0.95\textwidth,height=5cm]{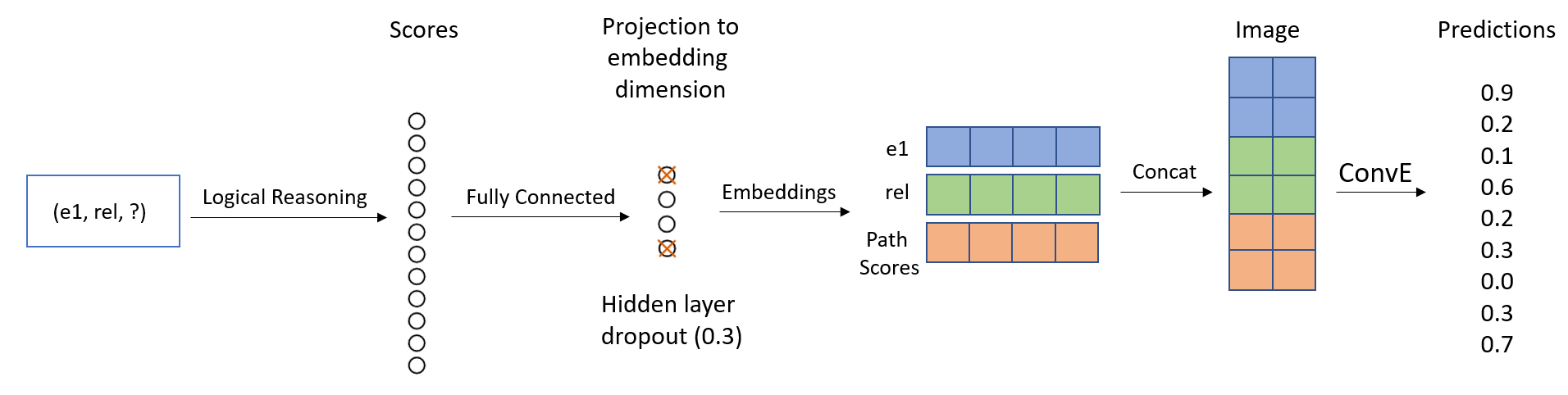}
  \caption{The modified version of the ConvE model designed to integrate the Neural and Logical models: When presented with a query, scores for various entities are computed using the Logical Reasoning module (steps 1 and 2). These calculated score vectors are then transformed into a k-dimensional space using a fully connected layer (step 3). Subsequently, the embeddings of the head entity, relation, and scores obtained from the logical model are reshaped and concatenated (steps 4 and 5). The resulting matrix serves as the input to the ConvE model, as discussed in Section \ref{sec:preliminaries}, to determine the final predictions. }
  \label{fig:fig4}
\end{figure}

\textbf{Using Neural Model for Inference Engine } 
In the second part of this section, we will delve into implementing the Reasoning Module in Figure 1 using neural models. The purpose of having a neural model is to have a model capable of training end-to-end combining the results of the neural and the logical model. Implementing the mentioned idea in the previous part using the neural model is a bit more challenging and different from using an algorithm. 

The central concept of the Reasoning Module is to merge the results of both neural and logical models through an Inference Engine. However, implementing the Reasoning Module with a neural approach presents a challenge. Unlike the algorithmic method, we cannot directly receive results from both the neural and logical models and use a neural Inference Engine to combine them. Our experiments have revealed that employing a neural Inference Engine to combine scores from both models hinders the model's learning ability. To address this challenge, we propose combining the scores from the logical model with the initial layers of the neural model. To put this idea into action, we modified the architecture of the ConvE model, as illustrated in Figure \ref{fig:fig4}. Illustrated in this figure, the revised ConvE model combines the neural and logical models. This approach involves concatenating the scores derived from the logical model with the embeddings of the head entity and the relation. Subsequently, this composite matrix is fed to the layers of the ConvE model. The modified score function of the ConvE model is defined below:

\begin{equation}
\label{equ:conve2}
f(vec(f([\overline{e_{h}}; \overline{r_r}, \overline{s_p}] * \omega)) W)e_t
\end{equation}

Equation \ref{equ:conve2} differs from the ConvE equation presented in Equation \ref{equ:conve} in that the input matrix to the convolutional layer consists of the concatenation of the head entity ($e_h$), relation ($r_r$), and the embedding derived from the scores of the logical model for the given query ($s_p$). The remainder of the architecture closely resembles the ConvE model. 

To calculate $s_p$, the process begins by providing a query in the format of (e1, r, ?) to the logical model. This query is used to obtain the top n answer entities, each associated with their respective scores (with reported results for n set to 10). Once the answer entities are obtained, a vector is created, sized according to the total number of entities. For each entity, if it is one of the answer entities, the corresponding element in the vector is set to the score computed by the logical model; otherwise, it is set to zero. This vector serves as the input to a fully connected layer, transforming it into a vector with the same dimensions as the embedding of entities and relations. This transformed vector is then used in Equation \ref{equ:conve2} as $s_p$.

In our proposed neural model, which combines both neural and logical models, the process begins by calculating a vector score for each query using the logical model. This vector score subsequently becomes one of the inputs for the neural model. It's worth noting that, in our neural model, the Inference Engine is not treated as a separate module, as is was proposed before in the algorithm idea. Instead, it is integrated with the neural model. What sets our model apart from previous ones is that we do not embed rules; rather, we utilize the results derived from them.

\subsection{Explanation Generation Module}
The Explanation Generation Module serves the purpose of generating natural language explanations for each entity present in the final answer set. To achieve this, the module follows a systematic process. Initially, it extracts the rules from the set of filtered rules that are capable of concluding the specific entity in question.

Subsequently, each of these extracted rules is passed through the Sentence Convertor module, as described in detail earlier. The Sentence Convertor transforms the logical rules into coherent sentences in natural language. This step allows the module to bridge the gap between symbolic reasoning and human-understandable explanations.

By leveraging this process, the Explanation Generation Module is capable of providing meaningful and contextually relevant explanations for each entity present in the final answer set. These explanations facilitate a deeper understanding of the reasoning behind the entity's inclusion in the answer set, making the overall results more interpretable and valuable to users.

\begin{table}[]
\captionof{table}{This table shows the results of reasoning on the FB15k-237 dataset. For the AnyBurl and ConvE models, the results are obtained through training, while for the other models, the results are extracted from the RuleGuider article \cite{lei-etal-2020-learning}. The best results are highlighted in \textbf{bold}. The second best results are underlined.}
\centering
\begin{adjustbox}{width=0.85\textwidth}
\begin{tabular}{l|llll}
\hline
                   & Hits@1         & Hits@5         & Hits@10        & MRR            \\ \hline
AnyBurl            & 26.67          & 42.77          & 51.54          & 34.75          \\ \hline
ConvE              & 34.0          & 54.7           & 62.3         & 43.5          \\
DistMult           & 32.4           & -              & 60.0           & 41.7           \\
ComplEx            & 33.7           & 54.0           & 62.4           & 43.5           \\
RotateE            & 34.1           & 53.2           & 62.2           & 43.5           \\ \hline
MINERVA            & 21.7           & -              & 45.6           & 29.3           \\
MultiHop (ConvE)   & 32.7           & -              & 56.4           & 40.7           \\
RuleGuider (ConvE) & 31.6           & 49.6           & 57.4           & 40.8           \\ \hline
FaSt-FLiP (Algorithm)               & \underline{34.76} & \underline{55.42} & \underline{63.27} & \underline{44.17} \\ \hline
FaSt-FLiP (Neural)               & \textbf{35.3} & \textbf{55.9} & \textbf{63.9} & \textbf{44.8} \\ \hline
\end{tabular}
\end{adjustbox}
\label{table:table1}
\end{table}

\section{Experiments}
\label{sec:experiments}

In this section, we begin by assessing our model's performance across different knowledge graphs, demonstrating its efficacy in the link prediction task through comparisons with logical, neural, and multi-hop models. Subsequently, we shift our focus to evaluating our model's ability to generate explanations, comparing it with logical and multi-hop models and highlighting its superior performance in producing more logical explanations. Additionally, we conduct experiments to showcase the effectiveness of different model components and explore the influences of hyperparameters. Furthermore, we delve into detailed analyses and ablation studies to gain a deeper understanding of our proposed approach.

\subsection{Experimental Setting}

\textbf{Datasets} The experiments in this article are conducted on the FB15k-237 Knowledge Graph \cite{toutanova2015observed}, which is a subset of the Freebase knowledge graph. In this subset, inverse relations that cause test leakages are omitted. Freebase itself is a knowledge graph extracted from Wikipedia, containing real-life information. Within this knowledge graph, entities encompass a wide range of real-life subjects, including humans, locations, professions, awards, and more. The relations within the FB15k-237 dataset encompass attributes such as gender, place of birth or death, winning an award, and so on. This dataset comprises 14,505 entities and 237 relations, with 272,115 triples in the training set, 17,535 in the evaluation set, and 20,467 in the test set.

\textbf{Baselines} In our comparative analysis, we assessed our model against three primary categories of link prediction models: logical, embedding, and multi-hop. For the logical model, we used the AnyBurl \cite{meilicke2019anytime} model as a baseline, which is one of the two key components of our framework. As for embedding models, we compared our results with ConvE \cite{dettmers2018convolutional}, which is another critical component of our model. Additional embedding models used for comparison include DistMult \cite{yang2014embedding}, ComplEx \cite{trouillon2016complex}, and RotateE \cite{sun2019rotate}. Finally, within the multi-hop category, we conducted comparisons with Minerva \cite{das2017go}, MultiHop (ConvE) \cite{lin-etal-2018-multi}, and RuleGuider (ConvE) \cite{lei-etal-2020-learning} models.

\textbf{Evaluation Protocols} In our evaluation, we adopted the filtered setting, consistent with previous studies \cite{lei-etal-2020-learning}. In this setting, when having a query in the format of (e1, r, ?), we generated an ordered list of entity answers from our model, adhering to the condition that for a given answer e2, the triple (e1, r, e2) should not be present in the train, validation, or test sets. Once we had this ordered list, we utilized it to compute Hits@N and Mean Reciprocal Rank (MRR) as evaluation metrics. For Hits@N, we checked whether the correct answer was among the top n in the list, while MRR involved summing the inverses of the positions of the correct entities and then averaging these values based on the number of triples.

\subsection{Results}
\textbf{Comparison for Link Prediction} Table \ref{table:table1} presents the results of the comparison on the FB15k-237 dataset. It is evident that our model surpasses models from all three categories. In particular, our model exhibits superior performance compared to AnyBurl and ConvE, its primary components, achieving improvements of approximately 13\% and 2\% in Hits@10, respectively. Furthermore, our model outperforms other neural models. Additionally, when compared to multi-hop models, our proposed model outperforms them by approximately 6\% in Hits@10. An interesting observation is that the RuleGuider model also combines AnyBurl rules and ConvE embeddings to create a reinforcement model. However, the results from this article indicate that the combination proposed here yields better results.

\textbf{Comparison for Explainablity} An integral component of our model is the Filter Rule module, and in this section, we aim to assess its influence on generating improved and more acceptable explanations. We have previously highlighted the issue of incorrect paths in logical and multi-hop models, emphasizing the importance of the Filter Rule module in mitigating this problem. Table 2 illustrates this concern. Within the table, we present two triples, along with some of the rules that could generate them. It is evident that for each of these two triples, there exist rules that are incorrect and would not be deemed logical by a human. The utilization of the Filter Rule module has effectively removed these erroneous rules, resulting in explanations that are more readily acceptable to human judgment.

\begin{figure}
  \centering
  \includegraphics[width=0.8\textwidth,height=8cm]{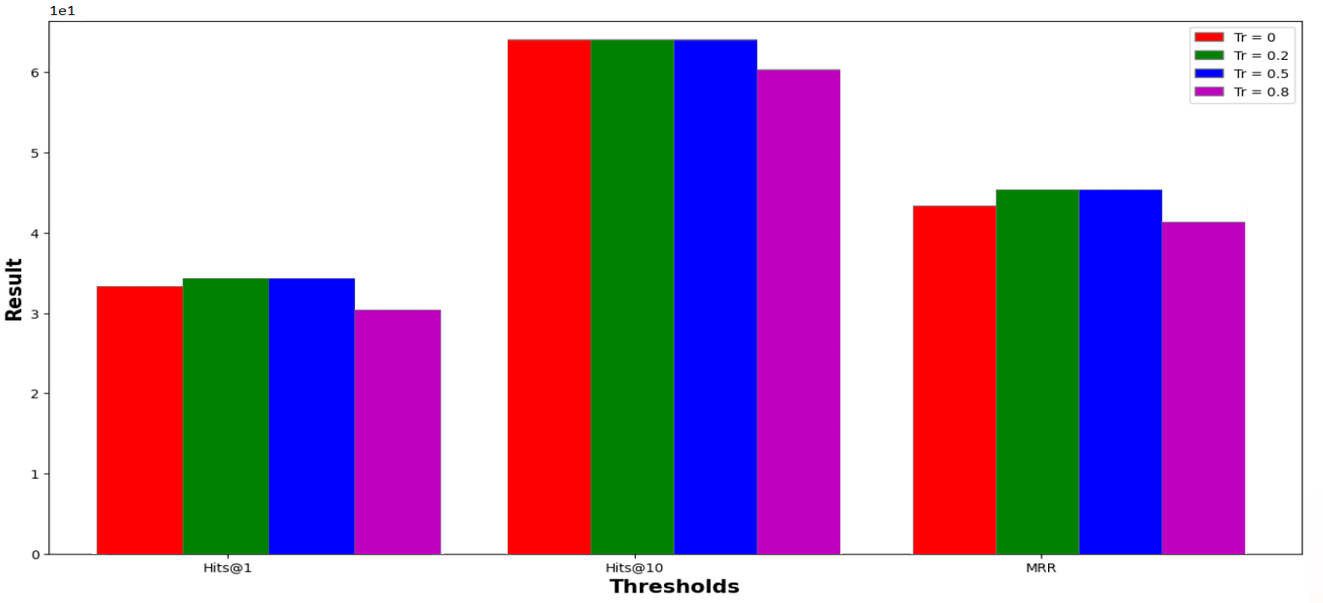}
  \caption{The effect of different thresholds in equation \ref{eqn:finalscore} on the results.}
  \label{fig:fig2}
\end{figure}

\begin{table}[]
\captionof{table}{Rules that are incorrect as an explanation are removed using the Filter Rule module.}
\begin{adjustbox}{width=1.1\textwidth}
\begin{tabular}{ll}
\textbf{Triple}                             & \textbf{Result Rules}                                                                                                                                                                                                                                                                                                                                                                                                                                                                     \\ \hline
$(/m/06sy4c, gender, /m/05zppz)$     & \begin{tabular}[c]{@{}l@{}}Correct rules:\\ $gender(X,Y) \leftarrow team(X,A), team(B,A), gender(B,Y)$ \\ \\ Incorrect rules:\\ $gender(X,Y) \leftarrow nationality(X,A), place\_lived/location(B,A), gender(B,Y)$ \end{tabular}                                                                                                                                                                                                                                                      \\ \hline
$(/m/0133sq, nationality, /m/02jx1)$ & \begin{tabular}[c]{@{}l@{}}Correct rules:\\ $nationality(X,Y) \leftarrow ethnicity(A,X),  ethnicity(A,B), place\_of\_birth(B,Y)$ \\ $nationality(X,Y) \leftarrow ethnicity(A,X), ethnicity/languages\_spoken(A,B), /language/countries\_spoken\_in(B,Y)$\\ \\ Incorrect rules:\\ $nationality(X,Y) \leftarrow /award/award\_winner(A,X), /award/award\_winner(A,B), nationality(B,Y)$ \\ $nationality(X,Y) \leftarrow profession(X,A), profession(B,A), nationality(B,Y)$ \end{tabular}
\end{tabular}
\end{adjustbox}
\label{table:table2}
\end{table}

\begin{figure}
  \centering
  \includegraphics[width=1.2\textwidth,height=8cm]{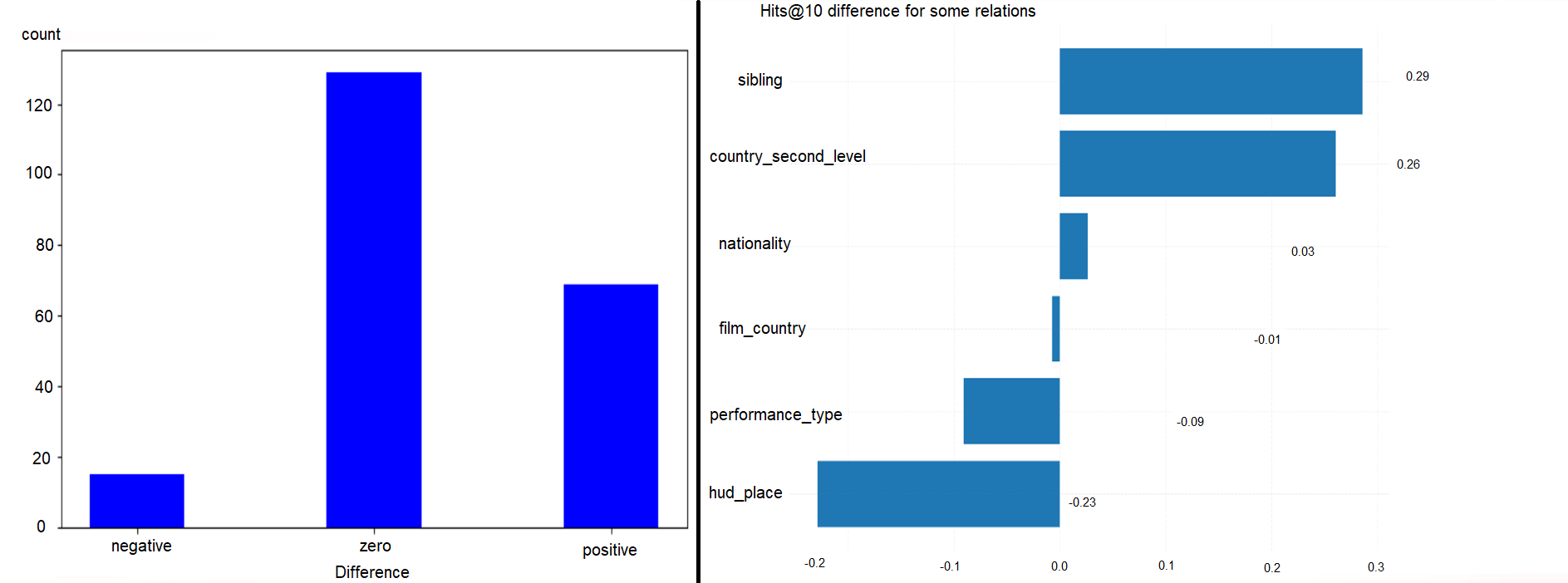}
  \caption{This Figure assesses the influence of our model in comparison to the ConvE model across triples associated with each relation. In the left graph, positive values indicate cases where our model outperforms ConvE, zero signifies similarity in performance, and negative values indicate ConvE's superiority. The right graph presents sample relations alongside their performance differences between our model and the ConvE model.}
  \label{fig:fig3}
\end{figure}

\subsection{Ablation Study}
Within this section, our objective is to conduct a comprehensive analysis of various aspects of our method. We will explore the influence of hyperparameters, assess how our neural-symbolic model influences different relations, and present the impact of the scores of the logical model on the training of the neural model.

\textbf{Impact of hyperparameters} We defined Equation \ref{eqn:finalscore} to compute the final score for each rule, incorporating a threshold within this equation. Rules with scores exceeding this threshold are retained, while those falling below it are omitted. Figure \ref{fig:fig2} illustrates the effects of varying thresholds on the final score. The figure reveals that thresholds of 0.2 and 0.5 yield slight improvements in results compared to a zero threshold. However, higher thresholds, such as 0.8, cause worse results. The important point here is that the Filter Rule module has been implemented on 10 out of the total 237 relations.

\textbf{Impact of the Neural-Symbolic model on different relations} An important question revolves around the impact of our proposed model on queries for different relations and whether it leads to overall improvements. To address this inquiry, we conducted an experiment to assess how our model influences queries related to various relations. In this experiment, we calculated Hits@10 for triples associated with each relation individually, both for our model and the ConvE model. Subsequently, we subtracted the Hits@10 results for each relation from the ConvE model from those of our model. The outcomes are graphically represented in Figure \ref{fig:fig3}. The left graph in the figure illustrates the number of relations where the subtraction result is positive (indicating our model's superiority), zero (suggesting equal performance), or negative (signifying ConvE's superior performance). On the right graph, specific relations with positive, zero, and negative results are highlighted. Notably, the figure reveals that for the majority of relations, our method did not have a significant impact. Nevertheless, the relations where our model had a positive impact outnumbered those with a negative impact by a factor of three. This observation suggests that our model succeeded in enhancing a larger number of relations.

\begin{figure}
  \centering
  \includegraphics[width=0.6\textwidth,height=5cm]{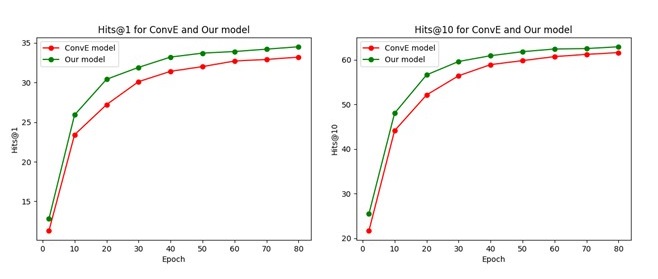}
  \caption{Comparison of the training procedures between our model and the ConvE model.}
  \label{fig:fig5}
\end{figure}

\textbf{Studying the training procedure of the neural inference engine model} In Section \ref{sec:proposed_model}, we introduced a neural model that incorporates the scores generated by the logical model as input during training. This approach allows the neural model to learn and adapt based on these scores, as well as the entity and relation embeddings. In this section, we delve into the training process of our proposed model and draw a comparison with the ConvE model. To conduct an assessment of both models, we subjected them to an 80-epoch training process. At regular intervals, specifically every 10 epochs, we computed the Hits@1 and Hits@10 metrics for both models on the test dataset. The training progress is visualized in Figure \ref{fig:fig5}. The training curves for our model and the ConvE model appear remarkably similar, highlighting that the combination of logical model scores doesn't significantly alter the training dynamics. However, a notable distinction emerges: at every epoch, our model consistently outperforms the ConvE model. This experiment's key insight is that providing logical model scores as input to the neural model effectively mimics the presence of an algorithmic Inference Engine. It doesn't revolutionize the learning process but substantially enhances the neural model's performance, resulting in consistently superior outcomes. It's worth highlighting an interesting observation: our model's performance after just 80 epochs surpasses that of the ConvE model after 1000 epochs. Specifically, our model achieved results of 62.9 for Hits@10 and 34.5 for Hits@1, compared to the ConvE model's 62.3 and 34.0, respectively. This finding underscores that the utilization of logical model scores has expedited the attainment of superior results, significantly reducing the required number of training epochs.

\section{Conclusion}
We introduced a novel Neural-Symbolic model named FaSt-FLiP to fuse both logical and neural models for enhancing the link prediction task. First, to address the issue of incorrect rules derived from the logical model, we put forward a semi-supervised approach. This method involves converting rules into sentences following a hypothesis-premise format. The resultant hypothesis and premise sentences serve as input to an NLI model, which provides a score indicating the correctness of each rule. Secondly, to integrate the neural model with the rules from the logical model, we propose an innovative idea that involves collecting results from both models and amalgamating them through an Inference Engine module. This module is implemented using both an algorithmic approach and a neural model. In the future, we intend to employ the method for removing incorrect rules for guiding multi-hop models in navigating more logical paths. Additionally, the idea of combining logical and neural models can be implemented in different ways and employing different models.

\bibliographystyle{unsrt}  
\bibliography{references}

\end{document}